\documentclass[letterpaper, 10pt, conference]{utils/ieeeconf}
\IEEEoverridecommandlockouts
\let\labelindent\relax

\usepackage{amsmath,amssymb,amsfonts}
\usepackage{amsthm}
\usepackage{algorithmic}
\usepackage[linesnumbered,ruled,vlined]{algorithm2e}
\usepackage{lipsum} 
\usepackage{bm}  
\usepackage{graphicx}
\usepackage{cite}
\usepackage{picinpar}
\usepackage{array}
\usepackage{stfloats}
\usepackage{url}
\usepackage{verbatim}
\usepackage{enumerate}
\usepackage{enumitem} 
\usepackage{cite}
\usepackage{caption}
\usepackage{makecell}
\usepackage{adjustbox}
\usepackage{multirow}
\usepackage{booktabs}
\usepackage{tabularx}
\usepackage[table]{xcolor}
\usepackage{mathrsfs}

\def\BibTeX{{\rm B\kern-.05em{\sc i\kern-.025em b}\kern-.08em
    T\kern-.1667em\lower.7ex\hbox{E}\kern-.125emX}}

\newtheorem{definition}{Definition}
\newtheorem{problem}{Problem}

\begin{document}

\setlength{\abovedisplayskip}{4pt}
\setlength{\belowdisplayskip}{4pt}
\setlength{\abovedisplayshortskip}{4pt}
\setlength{\belowdisplayshortskip}{4pt}

\title{\LARGE \bf Prescribed Performance Guided Safe Reinforcement Learning for Roundabout Navigation of Autonomous Vehicles}

\author{ Wenhan Li$^{\dag}$, Yangrui Zhang$^{\dag}$, Hang  Zhou, Peiran Liu, and Yiding Ji$^{*}$
\thanks{Authors are all with Robotics and Autonomous Systems Thrust, Systems Hub, the Hong Kong University of Science and Technology (Guangzhou), Guangzhou, Guangdong Province, China. (Emails: \{wli724, yzhang112, hzhou269, pliu868\}@connect.hkust-gz.edu.cn, jiyiding@hkust-gz.edu.cn).} 
\thanks{$^{\dag}$Contributed equally; $^{*}$Corresponding author.}
\thanks{Financial support: National Natural Science Foundation of China grants 62303389, 62373289; Guangdong Scientific Research Platform and Project Scheme grant 2024KTSCX039 and Youth Talent Support Program of Guangdong Association for Science and Technology grant SKXRC2025463.}
}

\maketitle

\begin{abstract}

Connected and Autonomous Vehicles (CAVs) face significant challenges in roundabout navigation due to dense traffic interactions and unpredictable agent behaviors, necessitating high levels of adaptability and safety. While the combination of Reinforcement Learning (RL) and Model Predictive Control (MPC) offers a promising approach, existing frameworks suffer from a structural mismatch where pointwise RL decisions fail to align with MPC's trajectory level optimization, and often lack rigorous safety guarantees. This work proposes RLTrajGen-PPMPC-CBF, a hierarchical framework for safe, smooth and comfort roundabout navigation, which regulates the coupling between RL-based planning and MPC-based control. We employ Prescribed Performance Constraints (PPC) to enforce a gradual, stable convergence between reference trajectories and control outputs, ensuring MPC tracking performance remains consistent with intended behavior from RL. Additionally, a Control Barrier Function (CBF) layer provides real-time safety corrections that are fed back into the RL process. Simulation results in the MetaDrive environment demonstrate that our method achieves a 6.25\% increase in average success rate and a 65.87\% reduction in Motion Sickness Dose Value (MSDV) compared to the optimal baseline, meanwhile accelerating training convergence. 

\end{abstract}


\section{Introduction}
The advancement of Intelligent Transportation Systems (ITS) has accelerated the deployment of Connected and Autonomous Vehicles (CAVs), enabling them to navigate diverse urban environments~\cite{ahmedTechnologyDevelopmentsImpacts2022,ranaConnectedAutonomousVehicles2023, huang2025differentiable, zhou2026physics}. Significant progress has been achieved in standard driving tasks, including lane keeping~\cite{dingAdaptiveMemoryEventTriggered2024,pachpolEnhancedLaneKeeping2024,dingIntelligentEventTriggered2025}, lane changing~\cite{zhangControlBarrierFunctionGuided2025,wangResearchAutonomousDriving2024}, and adaptive cruise control~\cite{heAdaptiveCruiseControl2019,xiaoComprehensiveReviewDevelopment2010}. However, roundabouts remain particularly challenging due to dense vehicle interactions and rapidly changing traffic conditions, where planning or control failures may compromise both safety and ride comfort~\cite{dingAdaptiveMemoryEventTriggered2024,nakayaExplorationDeepReinforcement2025}.

Current autonomous driving approaches mainly fall into learning-based and model-based methods. Reinforcement Learning (RL) has demonstrated strong capability in high-dimensional perception and decision-making~\cite{antonioMultiAgentDeepReinforcement2022,zhouDevelopmentEfficientDriving2020,liuCuriositydrivenReinforcementLearning2025,huCoMoCAVsCohesiveDecisionGuided2025a}. However, its model-free nature makes it difficult to explicitly incorporate vehicle dynamics and safety constraints~\cite{brunkeSafeLearningRobotics2022,wuRecentAdvancesReinforcement2024}. In contrast, Model Predictive Control (MPC) explicitly considers system dynamics and constraints, while Control Barrier Functions (CBFs) have been widely integrated into MPC to provide formal safety guarantees~\cite{seoSafetyCriticalControlNonaffine2022,xiaoSafeAutonomyControl2023,amesControlBarrierFunctions2019, zhang2026safety}. Nevertheless, model-based approaches generally exhibit limited adaptability in highly dynamic environments.

\begin{figure}[tp]
    \centering
    \includegraphics[width=\linewidth]{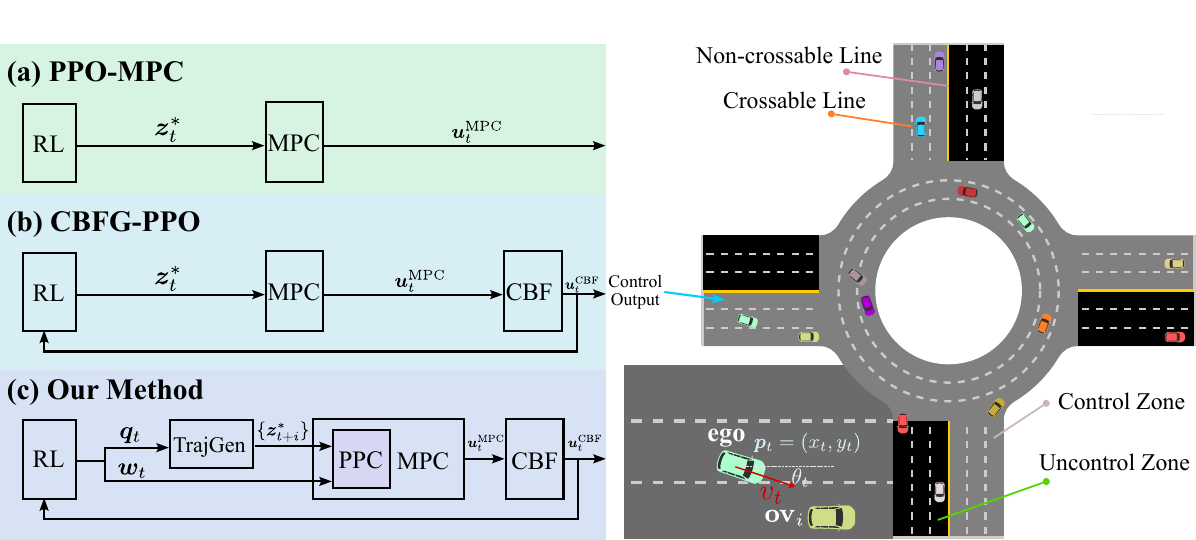}
    \caption{ Comparison between the proposed method (\textbf{bottom left}) and baseline approaches (\textbf{upper left}). We constrain the errors between RL and MPC outputs in prescribed performance bounds to improve trajectory planning performance. }
    \vspace{-15pt}
    \label{fig:roundabout}
\end{figure}

To combine the adaptability of RL with the capability to enforce safety and other constraints of MPC, hybrid RL-MPC frameworks have recently attracted increasing attention~\cite{reiterSynthesisModelPredictive2025a,zhouEnhancePlanningPhysicsinformed2024,oliveiradasilvaIntegratingReinforcementLearning2025a,lubarsCombiningReinforcementLearning2021}. Oliveira et al.~\cite{oliveiradasilvaIntegratingReinforcementLearning2025a} employed RL to determine discrete decision variables to accelerate MPC optimization, while Lubars et al.~\cite{lubarsCombiningReinforcementLearning2021} combined DDPG with MPC, where MPC served as a safety backup for potentially unsafe RL actions. Zhang et al.~\cite{zhangControlBarrierFunctionGuided2025} further incorporated a CBF safety filter into an RL-MPC framework (Fig. \ref{fig:roundabout}(b)). Nevertheless, these methods still suffer from a structural mismatch: RL produces pointwise decisions whereas MPC optimizes trajectories, limiting MPC's ability to accurately realize the planning intent and resulting in less smooth trajectories in complex driving scenarios.
Zhang et al.~\cite{zhangControlBarrierFunctionGuided2025} introduced a CBFG-PPO framework (Fig. \ref{fig:roundabout} (b)), where RL provides target states for MPC, and CBF filters unsafe actions and feeds corrections back to RL.
Nevertheless, RL produces pointwise decisions while MPC optimizes over trajectories in this framework, which results in structural mismatch and limits MPC’s ability to accurately realize RL intent, producing less smooth trajectories for CAVs in complex environments.


To address the mentioned issues, we propose a novel RLTrajGen-PPMPC-CBF framework, as illustrated in Fig. \ref{fig:roundabout} (c), and validate its effectiveness in a roundabout scenario. The proposed framework consists of three key components: an RL-based Trajectory Generator (RLTrajGen) that facilitates high-level decision making by producing reference trajectories, a prescribed performance MPC (PPMPC) that dynamically adjusts trajectory tracking within a predefined performance envelope, and a CBF-based safety filter that corrects unsafe behaviors. Specifically, we first introduce a trajectory generator that transforms the pointwise outputs of RL into reference trajectories for MPC tracking. In addition, we incorporate a prescribed performance constraint (PPC) mechanism to progressively reduce the tracking error, enabling MPC to more accurately follow the trajectories generated by RL. This design improves the consistency between planning and control, facilitating more stable policy learning and enhanced navigation performance. The main contributions of this work are summarized below:

\begin{enumerate}[leftmargin=*]
    \item We design a trajectory generation mechanism that converts single-step outputs from RL into trajectories with adaptive parameters for MPC tracking, thereby effectively resolving the structural mismatch between RL and MPC and significantly accelerating the convergence of training.
    \item We utilize PPC to regulate the coupling between RL-based planning and MPC-based control, which is the first one in RL-MPC hierarchical framework to our best knowledge. Our method facilitates the progressive reduction of trajectory tracking errors between the planning and control layers, thus balancing the adaptability to complex scenarios and the safety for autonomous systems, and improving ride comfort by producing smoother trajectories.
\end{enumerate}

The remaining sections are organized as follows. Section~\ref{sec:preliminaries} introduces the dynamic model of CAVs, prescribed performance constraints, control barrier functions, and formulates the safe navigation problem. Section~\ref{sec:method} presents the RLTrajGen-PPMPC-CBF framework. Section~\ref{sec:experiments} includes simulation results and compares our approach with baselines. Section~\ref{sec:conclusion} concludes the paper and outlines future directions.

\section{Preliminaries and Problem Formulation}\label{sec:preliminaries}
\subsection{System Model} \label{pre:sys-model}
The discrete-time dynamics of a CAV is modeled as
\begin{equation}
    \bm{x}_{k+1} = f(\bm{x}_k, \bm{u}_k)
    \label{eq:dynamic-eq}
\end{equation}
where the state $\bm{x}_k = [x_k, y_k, v_k, \theta_k]^T$ includes longitudinal and lateral positions, speed, and heading, respectively. The control input $\bm{u}_k = [a_k, \delta_k]^T$ comprises acceleration and steering. For the sake of discussion, we adopt the kinematic bicycle model to discretize the dynamics:
\begin{equation}
    \begin{array}{c}
    \begin{bmatrix} x_{k+1} \\ y_{k+1} \\ v_{k+1} \\ \theta_{k+1} \end{bmatrix}
    \end{array} = 
    \begin{array}{c}
    \begin{bmatrix}
        x_k + v_k\cos(\theta_k+\beta_k)\cdot T_s \\
        y_k + v_k\sin(\theta_k+\beta_k)\cdot T_s \\
        v_k + a_k\cdot T_s \\
        \theta_k + \dfrac{v_k\sin\beta_k}{l_r}\cdot T_s
    \end{bmatrix}
    \end{array}
    \label{eq:discrete-dynamic-evaluate-eq}
\end{equation}
where $T_s$ is the sampling period, $l_f$ and $l_r$ represent the distances from the front and rear wheels to the center of mass, and $\beta_k$ is the slip angle defined as $\beta_k=\arctan(\tan\delta_k\cdot l_r/(l_f+l_r))$. This model allows for iterative prediction of future vehicle states.


\subsection{Prescribed Performance Constraints}\label{pre:ppc}
To ensure asymptotic error convergence, the tracking error $e(t)$ can be constrained within a shrinking bound:
\begin{equation}
    -\sigma_L p(t) < e(t) < \sigma_R p(t)
    \label{eq:ppc-constraints}
\end{equation}
where the performance function $p(t)$ satisfies: 1) it is smooth, strictly positive, and monotonically decreasing; 2) $\underset{t\to0}{\lim}p(t)=p_0$, $\underset{t\to\infty}{\lim}p(t)=p_\infty$, with $p_0>p_\infty>0$. From \cite{buPrescribedPerformanceControl2023}, we utilize the form $p(t) = (p_0 - p_\infty)\mathrm{e}^{-\iota t} + p_\infty$, with decay rate $\iota \in [0,1]$. To incorporate PPC into controller design, the constrained error is transformed into an unconstrained one via $e(t) = \Xi(\zeta)p(t)$, where $\Xi(\zeta)$ is smooth, strictly increasing, bounded as $\Xi(\zeta) \in (-\sigma_L, \sigma_R)$ and given by:
\begin{equation}
    \Xi(\zeta) = \frac{\sigma_R \mathrm{e}^\zeta - \sigma_L \mathrm{e}^{-\zeta}}{\mathrm{e}^\zeta + \mathrm{e}^{-\zeta}}, \quad 
    \zeta = \frac{1}{2}\ln\frac{e(t)/p(t)+\sigma_L}{\sigma_R - e(t)/p(t)}.
    \label{eq:error-transformation-function}
\end{equation}

\subsection{Control Barrier Functions} \label{pre:cbf}
For system dynamics~\eqref{eq:dynamic-eq}, a superlevel safe set $\mathcal{C}$ is induced by a continuously differentiable function $h:\mathbb{R}^n\to\mathbb{R}$ as:
\begin{equation}
    \begin{aligned}
        \mathcal{C}               &= \{\bm{x}\in\mathbb{R}^n \mid h(\bm{x}) \ge 0\}, \\ 
        \partial \mathcal{C}      &= \{\bm{x}\in\mathbb{R}^n \mid h(\bm{x}) = 0\}, \\ 
        \mathrm{Int}(\mathcal{C}) &= \{\bm{x}\in\mathbb{R}^n \mid h(\bm{x}) > 0\}.
    \end{aligned}
    \label{safe-set-defination}
\end{equation}
And $h(\cdot)$ is defined as a discrete-time CBF~\cite{agrawalDiscreteControlBarrier2017} if it satisfies
\begin{equation}
    h(\bm{x}_{k+1}) - (1-\gamma_k)h(\bm{x}_k) \ge 0, \quad \gamma_k \in (0,1]
    \label{eq:cbf-condition-final}
\end{equation}
which guarantees the forward invariance of the safe set $\mathcal{C}$.  
By designing the control input $\bm{u}_k$ to satisfy~\eqref{eq:cbf-condition-final}, the system remains within $\mathcal{C}$ for all $t>0$ to maintain safe operation.

\subsection{Problem Formulation} \label{pre:prob-form}
A vehicle navigates in a two-dimensional roundabout (see Fig.~\ref{fig:roundabout}), where other vehicles appear randomly. Let $I^{\mathrm{ov}}=\{i \mid i\in\mathbb{Z}_{\ge0}\}$ represent the index set of vehicles within the control zone. When a new vehicle enters, the smallest unused index $i_{\mathrm{new}}=\min\{i\in\mathbb{Z}_{\ge0} \mid i\notin I^{\mathrm{ov}}\}$ is assigned. This index is subsequently removed when the vehicle exits, ensuring a dynamic mapping between indices and vehicles. To achieve safe and efficient roundabout navigation, the ego vehicle is required to optimize three objectives: travel time, control effort, and safety, which are later incorporated into the RL reward design and MPC optimization formulations.

\begin{definition}[Travel Time]\label{def:minitime}
The travel time for the ego vehicle to reach the destination is defined as a positive stage objective function $r_1(\bm{x})$ and should be minimized:
\begin{equation}
    J_1 = t_T - t_0 = \int_{t_0}^{+\infty} \delta(\bm{p}_\tau)\,\mathrm{d}\tau
    = \Phi\left[\int_{t_0}^{+\infty} r_1(\bm{x}_\tau)\,\mathrm{d}\tau\right]
    \label{cost-function-time}
\end{equation}
where $\bm{p}_\tau=(x_\tau,y_\tau)$ is the ego vehicle position, $\delta(\bm{p}_\tau)$ indicates whether the destination set $\mathcal{X}_T$ has been reached, and $\Phi(\cdot)$ is a strictly increasing mapping.   
\end{definition}


\begin{figure*}[t]
    \centering
    \includegraphics[width=\linewidth]{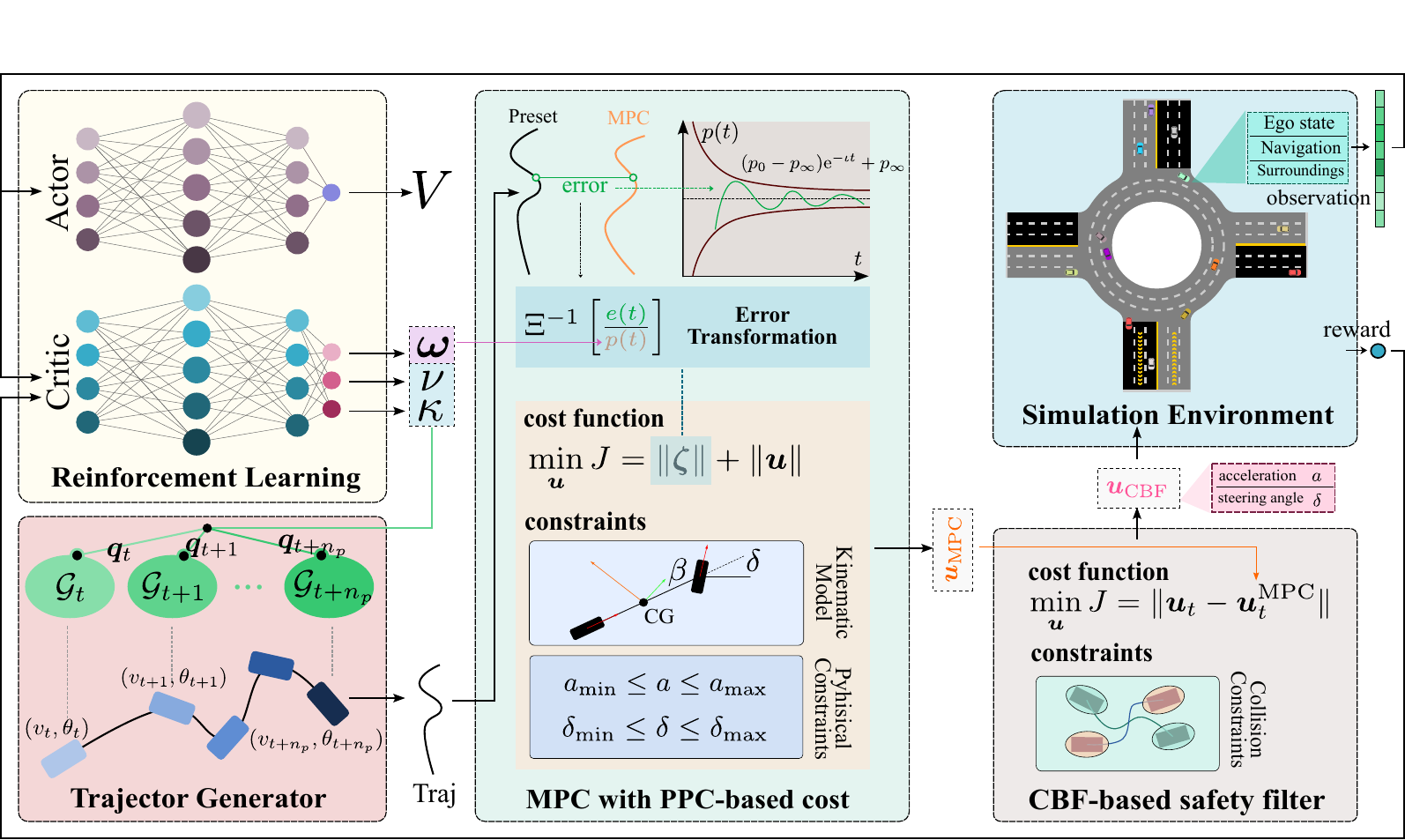}
    \caption{Overview of RLTrajGen-PPMPC-CBF framework. The RL module (top left) maps observations from the environment (top right) to actions $(\bm{q}_t, \bm{w}_t)$. The action $\bm{q}_t$ is processed by the trajectory generator (bottom left) to construct Trajectory Generator $\mathcal{G}_t$ and produce a prediction sequence $\bm{z}_t^*$ of length $n_p$, which is then used to solve an MPC problem yielding $u^{\mathrm{MPC}_t}$. This control input is refined by the CBF module (bottom right) to obtain $u^{\mathrm{CBF}_t}$, which is applied to the environment to generate rewards for RL training.}
    \vspace{-15pt}
    \label{fig:framework}
\end{figure*}

\begin{definition}[Energy Consumption]
Given a positive control cost function $C(\cdot)$, minimize the accumulated energy cost:
\begin{equation}
    J_2 =\int_{t_0}^{t_T} C(\bm{u}_\tau)\,\mathrm{d}\tau
    \label{cost-function-energy}
\end{equation}
\end{definition}


\begin{definition}[Safety Penalty]
Each vehicle should maintain a minimum safe distance $\delta$ from its surrounding vehicles, i.e., $\mathrm{dist}(\bm{p}_t,\bm{p}^{\mathrm{ov}_i}_t)\ge\delta$. Safety violations are quantified by a positive function $r_2(\bm{p},\bm{p}^{\mathrm{ov}})$ and the cumulative penalty is
\begin{equation}
    J_3 = \int_{t_0}^{t_T} r_2(\bm{p}_\tau,\bm{p}^{\mathrm{ov}_i}_\tau)\,\mathrm{d}\tau,
    \label{cost-function-safety-penal}
\end{equation}
where larger $r_2(\cdot)$ corresponds to more severe violations.
\end{definition}


\begin{problem}[Roundabout Navigation of CAVs]\label{prob:roundvav}
Given a CAV modeled by~\eqref{eq:dynamic-eq}, design the optimal control law $\bm{u}$ for safe navigation of the CAV and minimize the objective function:
\begin{equation}\label{final-problem}
    \begin{aligned}
    &\min_{\bm{x},\bm{u}}\ 
    \int_{t_0}^{+\infty}
    \Big[
    r_1(\bm{x}_\tau) + \big(C(\bm{u}_\tau) + r_2(\bm{p}_\tau,\bm{p}^{\mathrm{ov}_i}_\tau)\big) \delta(\bm{p}_\tau)
    \Big]\,\mathrm{d}\tau \\
    &s.t.\left\{\begin{array}{l}
    \bm{x}_\tau \in \mathcal{C}_{\mathrm{safe}} \\
    \bm{u}_{\min}\le \bm{u}\le \bm{u}_{\max}
    \end{array}\right.
    \end{aligned}
\end{equation}
where $u_{\min}$, $u_{\max}$ denote the lower and upper bounds of $\bm{u}$, respectively, and $\mathcal{C}_{\mathrm{safe}}$ defined from~\eqref{safe-set-defination} denotes the safe set for collision avoidance of the vehicles.
\end{problem}


\section{hierarchical framework for safe navigation}\label{sec:method}
\subsection{Framework Overview}\label{med:overview}
The proposed framework consists of three coupled layers, as illustrated in Fig. \ref{fig:framework}: 
1) a high-level decision-making layer that specifies reference motion intentions through a learned policy; 
2) a low-level control layer that tracks the resulting reference trajectory and computes control inputs based on the dynamic model; and 
3) a safety mediation layer that corrects unsafe behaviors during execution.

In general, the framework follows a standard reinforcement learning training and evaluation pipeline. 
At each time step, the high-level agent produces actions that generate a reference state sequence $\bm{z}^*_t$ of length $n_p$:
\begin{equation}
    \bm{z}^*_t = \{\bm{z}^*_{t+i}\} = \{(v^*_{t+i}, \theta^*_{t+i}) \mid i=0,1,\cdots,n_p-1\}
    \label{eq:state-sequence}
\end{equation}
where $\bm{z}^*_t$ represents the desired future motion of the ego vehicle. 
The low-level controller is responsible for tracking this reference trajectory, and the corresponding tracking error $e$ incorporated into the cost function MPC is given as:
\begin{equation}
    e = \mathrm{dist}(\bm{z}_t, \bm{z}^*_t)
    \label{eq:mpc-rl-error}
\end{equation}
where $\bm{z}_t$ denotes the trajectory generated by MPC and $\mathrm{dist}(\cdot)$ measures the deviation from the reference. 
After any potentially unsafe action output from the MPC is corrected using a CBF-based filter, the resulting control input $\bm{u}_t = [a_t, \delta_t]^T$ is applied to the simulated vehicle. This interaction with the environment yields a reward, thereby completing one training iteration.

\begin{table}[tp]
    \centering
    \caption{Rewards Composition}\label{tab:rewards}
    \begin{tabularx}{\linewidth}{ l c c }
\toprule
Reward                   & Explanation & Definition \\
\midrule
$\bm{r_1}$               &             &            \\
$r_{\mathrm{dest}}$      & Progress to destination & $\Vert \bm{l}_t - \bm{l}_{t-1} \Vert$            \\
$r_{\mathrm{speed}}$     & Speed regulation        & $v_t / v_{\max}$                       \\
$r_{\mathrm{steer}}$     & Steering change penalty & $-\Vert \theta_t - \theta_{t-1} \Vert$ \\
$r_{\mathrm{terminal}}$  & Terminal reward         & $10$ if arrived, else $0$              \\
\midrule
$\bm{r_2}$               &             &            \\
$r_{\mathrm{crash}}$     & Crash penalty           & $-5$ if crash, else $0$                \\
\bottomrule
\end{tabularx}

    \vspace{-15pt} 
\end{table}

\subsection{RL-based Trajectory Generator}\label{med:rl}
The surrounding environment is perceived through onboard sensors, such as LiDAR, and encoded into a $259$-dimensional observation vector. This vector contains various parameters including the vehicle speed $v_t$, steering angle $\delta_t$, previous throttle/brake command $u_{t-1}$, previous steering angle $\delta_{t-1}$, yaw rate $r_t$, heading difference to the lane $\Delta \psi$, distances to the left and right lane boundaries $d_L$ and $d_R$, lateral lane position $p_{lat}$, projected distances along the lane heading $d_{\parallel}$, projected lateral distances $d_{\perp}$, lane curvature radius $R$, turn direction flags $C$, lane angles $\theta$, and surrounding LiDAR features $\bm{s}^{sur}_t$.

\begin{figure}[tp]
    \centering
    \includegraphics[width=0.95\linewidth]{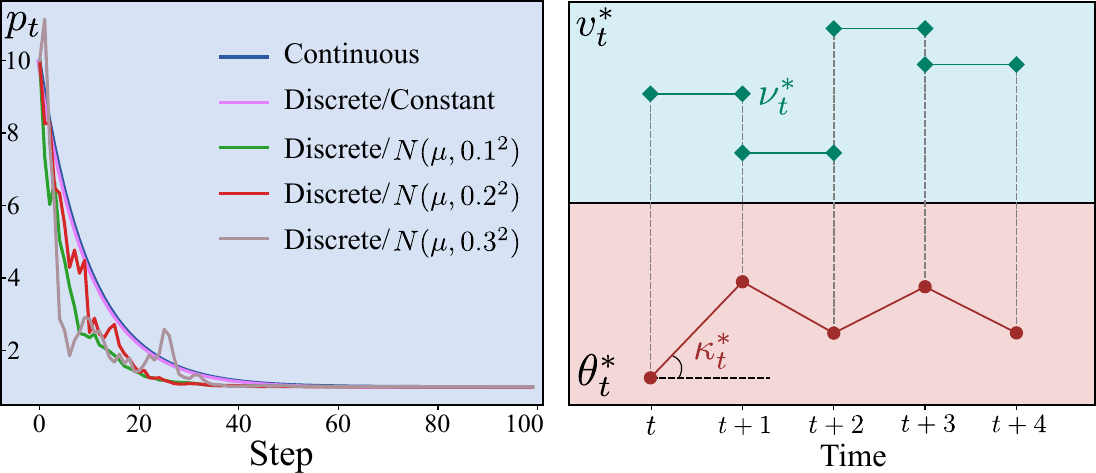}
    \caption{(\textbf{left}) Discrete-time performance verification for five cases: continuous-time baseline, constant weight($\mathrm{e}^{-\iota\Delta t}$), and random weights from normal distributions $\mathcal{N}\sim(\mathrm{e}^{-\iota\Delta t}, \sigma^2)$ where $\sigma=0.1/0.2/0.3$. (\textbf{right}) Schematic of the reference Trajectory Generator parameterized by $(\nu_t^*, \kappa_t^*)$.}
    \vspace{-5pt}
    \label{fig:traj-ppc}
\end{figure}

The action space includes adjustable parameters for a generated reference trajectory and its prescribed performance tracking weights. Specifically, the action is defined as $\bm{a}_t = (\bm{q}_t, \bm{w}_t)$. 
$\bm{w}_t = [w_t^v, w_t^\theta] \in \mathbb{R}^2$ is a weight vector that adjusts the strictness of the prescribed  trajectory tracking performance in the MPC.
$\bm{q}_t = [\nu_t^*, \kappa_t^*] \in \mathbb{R}^2$ specifies the structure of a Trajectory Generator $\mathcal{G}_t(\bm{x}_t;\bm{q}_t):\mathbb{R}^4\to\mathbb{R}^{2\times n_p}$ and helps to generate the reference trajectory by the following state transition function:
\begin{equation}
    \left\{
    \begin{array}{l}
    v_{t+i}^* = \nu_t^*, \, i=0,\cdots,n_p-1, \\
    \theta_{t+i}^* = \theta_{t+i-1}^* + v_{t+i-1}^*\kappa_t^*\Delta t, \, i=1,\cdots,n_p-1, \\
    \theta_t^* = \theta_t.
    \end{array}
    \right.
    \label{eq:traj-generator}
\end{equation}
Fig. \ref{fig:traj-ppc} illustrates a segment of the resulting reference trajectory. The velocity profile is piecewise constant, while the heading angle evolves as a continuous piecewise linear function. By modifying the Trajectory Generator, different reference trajectory shapes can be obtained.

Based on the optimal control problem defined in \eqref{final-problem}, two reward terms are designed: $r_1$ for goal-reaching behavior and $r_2$ for collision avoidance, which are presented in Table~\ref{tab:rewards} in detail. 
Specifically, $r_1$ is composed of four components as $r_1 = r_{\mathrm{dest}} + r_{\mathrm{speed}} + r_{\mathrm{steer}} + r_{\mathrm{terminal}}$. Here, $r_{\mathrm{dest}} = \Vert \bm{l}_t - \bm{l}_{t-1} \Vert$ measures progress toward the goal, $r_{\mathrm{speed}} = v_t / v_{\max}$ encourages efficient motion, $r_{\mathrm{steer}} = -\Vert \theta_t - \theta_{t-1} \Vert$ penalizes abrupt steering changes, and $r_{\mathrm{terminal}} = 10$ is assigned upon successful arrival. 
Meanwhile, the collision-related reward is defined as $r_2 = r_{\mathrm{crash}}$, where $r_{\mathrm{crash}} = -5$ in the event of a collision and $0$ otherwise.

\begin{figure}[tp]
    \centering
    \includegraphics[width=0.95\linewidth]{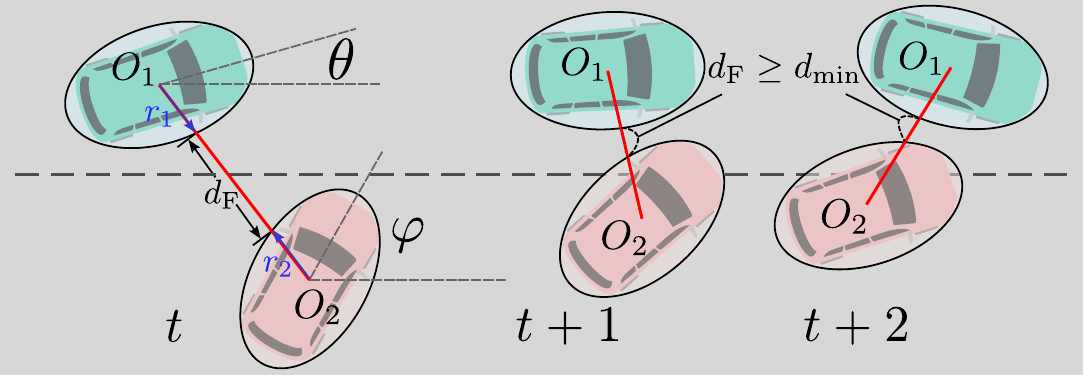}
    \caption{The establishment of collision avoidance diagrams.}
    \vspace{-15pt}
    \label{fig:cbf-dem}
\end{figure}

Proximal Policy Optimization (PPO) is adopted for policy learning. 
The policy network $\pi_\theta(\bm{a}_t|\bm{o}_t)$ generates actions from observations, while the value network $V_\omega(\bm{o}_t)$ estimates the state value. 
The Generalized Advantage Estimation (GAE) is computed as
\begin{equation}
    \hat{A}_t = \delta_t + \gamma\lambda(1-d_{t+1})\hat{A}_{t+1},
    \label{eq:gae-formulation}
\end{equation}
\begin{equation}
    \delta_t = r_t + \gamma(1 - d_{t+1})V_{t+1} - V_t
    \label{eq:caculate-delta}
\end{equation}
where $V_t = V_\omega(\bm{o}_t)$ represents the value function, $r_t$ denotes the reward, $d_t$ indicates episode termination, $\gamma$ is the discount factor, and $\lambda$ controls the bias--variance trade-off.

The loss functions of the two networks in PPO are defined:
\begin{equation}
    L^{\mathrm{CLIP}}(\theta) = \hat{\mathbb{E}}_t[\min(r_t(\theta)\hat{A}_t, \mathrm{clip}(r_t(\theta), 1-\varepsilon, 1+\varepsilon)\hat{A}_t)],
    \label{eq:policy-loss-function}
\end{equation}
\begin{equation}
    L(\omega) = \hat{\mathbb{E}}_t[(r_t + \gamma V_\omega(\bm{o}_{t+1}) - V_\omega(\bm{o}_t))^2]
    \label{eq:value-loss-function}
\end{equation}
where $\mathrm{clip}$ is a clipping function that constrains $r_t(\theta)$ to $[1-\varepsilon, 1+\varepsilon]$, and $r_t(\theta) := \pi_\theta(\bm{a}_t|\bm{o}_t) / \pi_{\theta_{\mathrm{old}}}(\bm{a}_t|\bm{o}_t) $. The final training objective is a weighted combination of the policy and value losses:
\begin{equation}
    L_{\mathrm{final}} = a_1 L^{\mathrm{CLIP}}(\theta) + a_2 L(\omega),\, a_i \in (0,1),\, i=1,2.
    \label{eq:loss-final}
\end{equation}

\subsection{MPC with PPC-based cost}\label{med:mpc-ppc}
Prescribed performance constraints (PPC) progressively shrinks the error bounds, ensuring that the error asymptotically converges to a specified value. This characteristic can be directly integrated into the construction of $ \mathrm{dist}(\bm{z}_t, \bm{z}^*_t)$ in \eqref{eq:mpc-rl-error}. Given that $ \mathrm{dist}(\bm{z}_t, \bm{z}^*_t)$ is evaluated in discrete time, the performance function in PPC must first be discretized.
Denoting the sampling interval as $\Delta t$ and defining a discrete-time performance function $p_k = p(k\Delta t)$, we obtain:
\begin{equation}
    p_{k+1} = p((k+1)\Delta t) = w p_k + (1 - w)p_{\infty}
    \label{eq:dicrete-time-equation-derivation}
\end{equation}
Following \cite{buDiscreteTimePrescribedPerformance2023, liuPrescribedPerformanceModelFree2019}, 
the parameter $w = \exp(-\iota\Delta t)$ is treated as time-varying, i.e., $w=w_k:=w(k)$. Then, \eqref{eq:dicrete-time-equation-derivation} can be rewritten as:
\begin{equation}
    p_{k+1} = w_k p_k + (1 - w_k)p_{\infty},\, k=0,1,\dots,+\infty
    \label{eq:dicrete-time-equation-derivation-final}
\end{equation}
The effectiveness of \eqref{eq:dicrete-time-equation-derivation-final} can be verified in Fig. \ref{fig:traj-ppc}. The error transformed by this discrete-time performance metric is used to define the distance function:
\begin{equation}
    \begin{aligned}
    \mathrm{dist}(\bm{z}_t, \bm{z}_t^*) 
    &= \mathrm{dist}([v_t,\theta_t],[v_t^*,\theta_t^*]) \\
    &= \left\Vert\big[\Xi^{-1}(\bar{e}_t^v), \Xi^{-1}(\bar{e}_t^\theta)\big]\right\Vert
    \end{aligned}
    \label{eq:dist-definetion}
\end{equation}
where the normalized errors $\bar{e}_t^v$, $\bar{e}_t^\theta$ are defined as
\begin{equation}
    \bar{e}_t^v = \frac{v_t - v_t^*}{p_\tau^v}, \,
    \bar{e}_t^\theta = \frac{\theta_t - \theta_t^*}{p_\tau^\theta}.
    \label{eq:barerror-explanation}
\end{equation}
Here, $p_\tau^v$ and $p_\tau^\theta$ are the discrete-time performance functions for $v$ and $\theta$, respectively. The two performance functions correspond to weights $\bm{w}_t = [w_t^v, w_t^\theta] \in \mathbb{R}^2$ introduced in Sec.\ref{med:rl}.
The PPC timer is reset at the beginning of each episode, and $\tau$ denotes the relative time within the episode. 

With the defined PPC-based distance function, the MPC problem is then formulated as
\begin{equation}
    \begin{aligned}
    &\underset{\bm{u}}{\min}\; J(\bm{z}_t^{t+n_p},\bm{u}_t^{t+n_p-1}) = \\
    &\sum_{i=0}^{n_p-1}
    \left[
    \underbrace{\mathrm{dist}(\bm{z}_{t+i}, \bm{z}_{t+i}^*)}_{\text{state error}}
    +
    \underbrace{\Vert \bm{u}_{t+i}\Vert}_{\text{control cost}}
    \right]
    +
    \underbrace{\mathrm{dist}(\bm{z}_{t+n_p}, \bm{z}_{t+n_p}^*)}_{\text{terminal cost}} \\
    &s.t.
    \left\{\begin{array}{l}
    \bm{x}_{t+i+1} = \hat{f}(\bm{x}_{t+i}, \bm{u}_{t+i}), \\
    \bm{z}_{t+i} = [\bm{0}_{2\times 2}, \bm{I}_{2\times 2}] \bm{x}_{t+i}, \\
    \bm{u}_{\min} \le \bm{u}_{t+i} \le \bm{u}_{\max}, \, i=0,\dots,n_p-1
    \end{array}\right.
    \end{aligned}
    \label{eq:mpc-formulation}
\end{equation}
where $\bm{0}_{2\times 2}$ and $\bm{I}_{2\times 2}$ denote the zero and identity matrices, respectively, and $\hat{f}$ follows the dynamics in \eqref{eq:discrete-dynamic-evaluate-eq}. 
Solving \eqref{eq:mpc-formulation} yields the optimal state and control sequences. 
We define $\bm{x}_t^{\mathrm{MPC}} = \bm{x}_{t+1}$, 
$\bm{z}_t^{\mathrm{MPC}} = [\bm{0}_{2\times 2}, \bm{I}_{2\times 2}] \bm{x}_t^{\mathrm{MPC}}$, 
and denote the applied control as $\bm{u}_t^{\mathrm{MPC}}$, 
which will be used in Sec.\ref{med:cbf}.

\subsection{CBF-based safety filter}\label{med:cbf}
CBF is employed to synthesize a safety filter to correct potentially unsafe behaviors generated by MPC. 
In a roundabout scenario, the primary safety objective is to avoid inter-vehicle collisions. 
To this end, each vehicle is approximated by an ellipse, which enables a quantitative formulation of collision-avoidance constraints. 
As illustrated in Fig. \ref{fig:cbf-dem}, a vehicle is first approximated by a rectangle and then enclosed by a circumscribed ellipse. 
Let $a$ and $b$ denote the semi-major and semi-minor axes, respectively. 
The equation of a non-rotated ellipse centered at $\bm{x}_o$ is $(\bm{x}-\bm{x}_o)^T \bm{D} (\bm{x}-\bm{x}_o)=1$, where $\bm{D}=\mathrm{diag}(1/a^2,\,1/b^2)$.

For an ellipse rotated by an angle $\theta$, a coordinate transformation is applied using the rotation matrix $\bm{R}_\theta=[\cos\theta,\sin\theta;-\sin\theta,\cos\theta]$, yielding
$(\bm{x}-\bm{x}_o)^T \bm{R}_\theta^T \bm{D} \bm{R}_\theta (\bm{x}-\bm{x}_o)=1$.
Then, let $\bm{Q}_\theta=\bm{R}_\theta^T \bm{D}\bm{R}_\theta$. 
To compute the distance between two ellipses, consider the line connecting their centers, along which a point is given by $\bm{x}=\bm{x}_o+\bm{e}t$, where $\bm{e}$ is the unit direction vector. 
Substituting into the ellipse equation yields $t=1/\sqrt{\bm{e}^T\bm{Q}_\theta\bm{e}}$. 
Accordingly, the distance between the boundaries of two ellipses is defined as
$d_{\mathrm{F}}(\bm{x})=\Vert \bm{e}_{o_{1,2}}\Vert-1/\sqrt{\bm{e}^T\bm{Q}_\theta\bm{e}}-1/\sqrt{\bm{e}^T\bm{Q}_\varphi\bm{e}}$,
where $\bm{e}_{o_{1,2}}=\bm{x}_{o_1}-\bm{x}_{o_2}$ and $\bm{e}=\bm{e}_{o_{1,2}}/\Vert \bm{e}_{o_{1,2}}\Vert$. 
The obstacle-avoidance function is then defined as $h(\bm{x}) = d_{\mathrm{F}}(\bm{x}) - d_{\min}$. 
For clarity, only one safety constraint is considered here. 
The resulting CBF optimization problem is formulated as:
\begin{equation}
    \begin{aligned}
    &\underset{\bm{u}}{\min}\, J(\bm{u}_t) = \Vert \bm{u}_t - \bm{u}_t^{\mathrm{MPC}}\Vert \\
    &s.t.\left\{\begin{array}{l}
    \bm{x}_{t+1} = \hat{f}(\bm{x}_t, \bm{u}_t),  \\
    h(\bm{x}_{t+1}) - (1 - \gamma)h(\bm{x}_t) \ge 0.
    \end{array}\right.
    \end{aligned}
    \label{eq:cbf-formulation}
\end{equation}
Solving \eqref{eq:cbf-formulation} yields $\bm{x}_{t+1}$ and $\bm{u}_t$, which are denoted as $\bm{x}_t^{\mathrm{CBF}} = \bm{x}_{t+1}$ and $\bm{u}_t^{\mathrm{CBF}} = \bm{u}_t$. 
The corresponding state representation is defined as $\bm{z}_t^{\mathrm{CBF}} = [\bm{0}_{2\times2}, \bm{I}_{2\times2}] \bm{x}_t^{\mathrm{CBF}}$.

Inspired by \cite{zhangControlBarrierFunctionGuided2025}, the policy loss term in \eqref{eq:policy-loss-function} is strategically modified to allow the reinforcement learning agent to learn CBF-induced corrective behavior:
\begin{equation}
    \begin{aligned}
    &L^{\mathrm{CLIP}}(\theta) = \\
    &\left\{\begin{array}{l}
    \mathbf{case\;1}\; \Vert\bm{z}_t^{\mathrm{CBF}}-\bm{z}_t^{\mathrm{MPC}}\Vert<\delta_{\bm{z}} \\
    \hat{\mathbb{E}}_t[\min(r_t(\theta)\hat{A}_t, \mathrm{clip}(r_t(\theta), 1-\varepsilon, 1+\varepsilon)\hat{A}_t)],\\
    \\
    \mathbf{case\;2}\; \Vert\bm{z}_t^{\mathrm{CBF}}-\bm{z}_t^{\mathrm{MPC}}\Vert \ge\delta_{\bm{z}} \\
    \hat{\mathbb{E}}_t[\bm{z}_t^{\mathrm{CBF}}-[\bm{I}_{2\times2},\bm{0}_{2\times1}]\bm{\mu}_t(\theta)].
    \end{array}\right.
    \end{aligned}
    \label{eq:policy-loss-function-modify}
\end{equation}
Here, $\delta_{\bm{z}}$ denotes the threshold beyond which the CBF intervention is considered significant. If that is the case, the original loss term is replaced by \textbf{case 2}, encouraging the policy mean $\bm{\mu}_t(\theta)= \mathbb{E}_{\bm{a}_t\sim \pi_\theta(\bm{a}_t|\bm{o}_t)}[\bm{a}_t] \in \mathbb{R}^4$ to better align with the CBF-corrected state.

\section{Experiments}\label{sec:experiments}

\subsection{Simulation Platform Setup}

\begin{figure}[tp]
    \centering
    \includegraphics[width=0.95\linewidth]{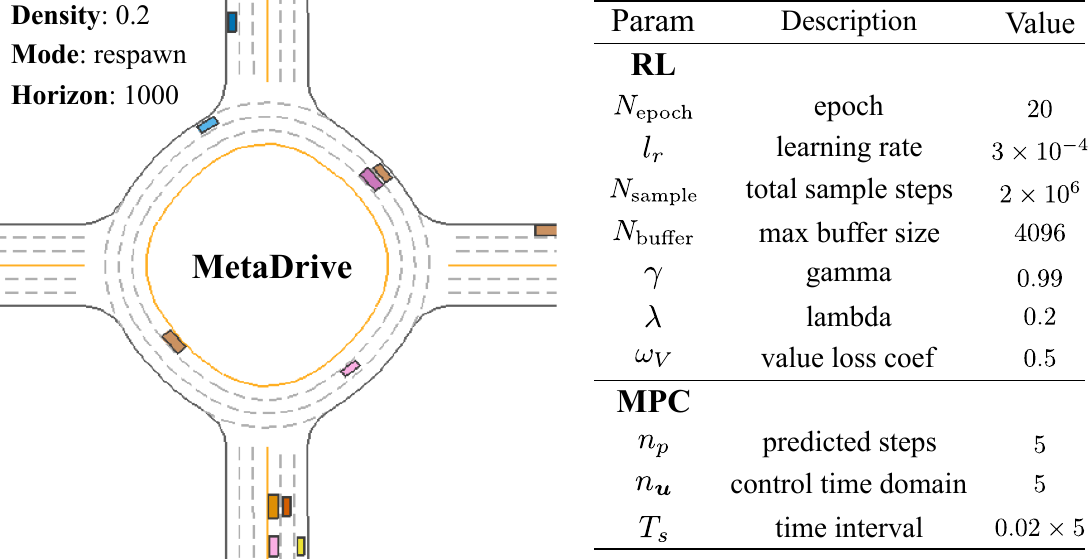}
    \caption{(\textbf{left}) Simulation environment diagram. (\textbf{right}) Training Hyperparameter Table.}
    \vspace{-10pt}
    \label{fig:metadrive-env}
\end{figure}

The roundabout simulation environment is built in MetaDrive and is illustrated in Fig. \ref{fig:metadrive-env}, with some key parameters specified as follows. The computer running reinforcement learning training is equipped with an AMD EPYC 7453 processor, with CUDA version 12.9 and Python version 3.11.4. During training, four parallel environments are used for sampling, followed by training on an NVIDIA RTX 4090 GPU. A fixed random seed is set at the initialization of each process to ensure the reproducibility of the experiments. The key hyperparameters of the framework listed in Fig. \ref{fig:metadrive-env} have been manually carefully tuned.

\subsection{Benchmark Comparison}

\begin{table}[tp]
    \centering
    \caption{Methods Overview}
    \begin{tabular}{m{1cm} m{1.2cm} m{5cm}}
    \toprule
    Index         & Name                    & Description        \\
    \midrule
    \textbf{Benchmark}     &                 &                    \\
    \textbf{M1}   & PPO-only                 & The RL policy directly outputs control commands $[a, \delta]$ to the simulator, following the official MetaDrive tutorial~\cite{liMetaDriveComposingDiverse2022} and serving as a standard baseline. \\
    \midrule
    \textbf{M2}   & PPO-MPC                  & RL provides high-level decisions, while MPC acts as the low-level controller, without additional coupling modules. \\
    \midrule
    \textbf{M3}   & CBFG-PPO                & A safety-augmented variant of \textbf{M1} that incorporates a CBF layer to enforce safety constraints and guide subsequent updates in the learning process~\cite{zhangControlBarrierFunctionGuided2025}. \\
    \midrule
    \textbf{Ablation}     &                 &                    \\
    \textbf{M4}   & RLTrajGen-MPC-CBF       & Replaces the RL module in \textbf{M3} with RLTrajGen to generate a reference trajectory for MPC tracking. \\
    \midrule
    \textbf{M5}   & RL-PPMPC-CBF            & Replaces the MPC module in \textbf{M3} with PPMPC, applying progressive penalties to tighten trajectory tracking errors. \\
    \midrule
    \textbf{M6}   & Our Method              & The full RLTrajGen-PPMPC-CBF framework, combining RL-based trajectory generation with progressively constrained MPC. \\
    \bottomrule
\end{tabular}
    \vspace{-10pt}
    \label{tab:methods}
\end{table}

\begin{figure}[t]
    \centering
    \includegraphics[width=\linewidth]{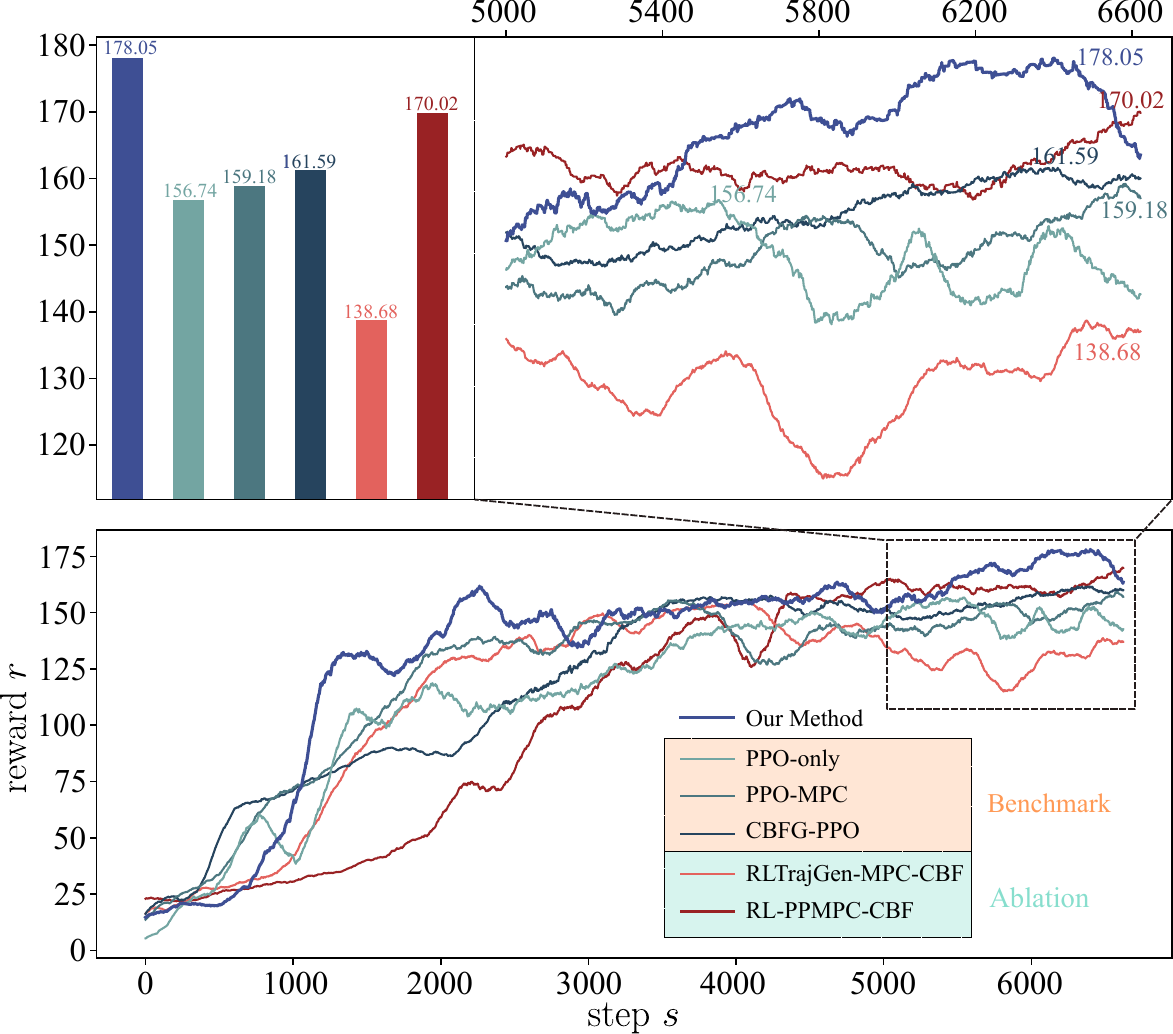}
    \caption{ AR evolution curves of all methods during training. }
    \vspace{-5pt}
    \label{fig:avg-reward}
    \vspace{-5pt}
\end{figure}

\begin{figure}[t]
    \centering
    \includegraphics[width=\linewidth]{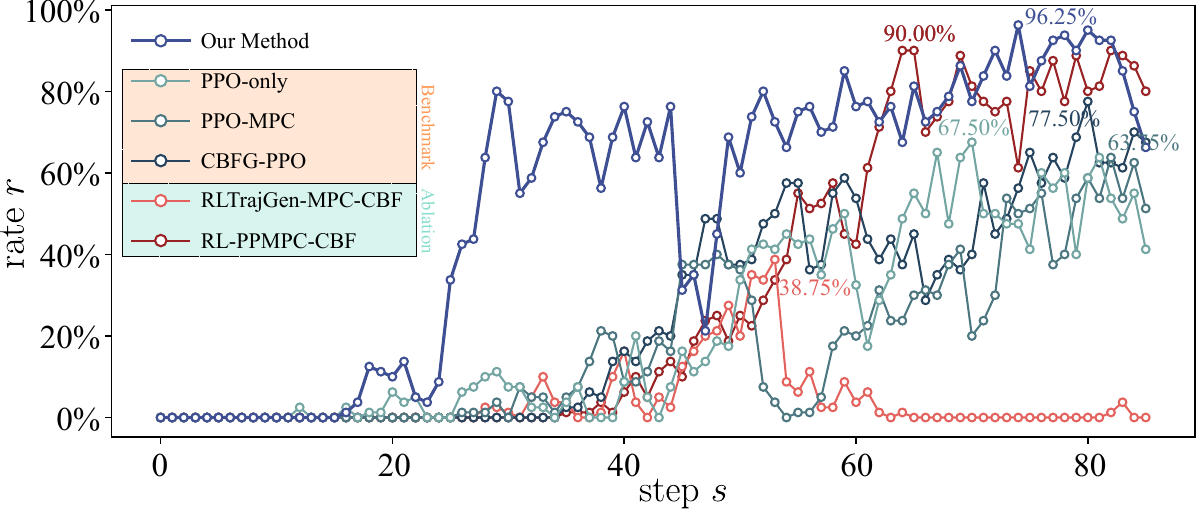}
    \caption{ ASR evolution curves of all methods during training. }
    \vspace{-15pt}
    \label{fig:success-rate}
    \vspace{-5pt}
\end{figure}

We select three baselines as the main comparison methods, namely PPO-only, PPO-MPC, and CBFG-PPO from~\cite{zhangControlBarrierFunctionGuided2025}. Detailed descriptions of these methods are provided in Table~\ref{tab:methods}. During training, a monitor is employed to record the average reward(AR) and average success rate(ASR) of each episode. The collected data are smoothed using a sliding window and are illustrated in Fig. \ref{fig:avg-reward} and Fig. \ref{fig:success-rate}. As shown in these figures, \textbf{M6} achieves the highest AR and ASR, reaching 178.05 and 96.25\%, respectively. These results outperform PPO-only (AR 156.74, ASR 67.50\%), PPO-MPC (AR 159.18, ASR 63.75\%), and CBFG-PPO (AR 161.59, ASR 77.50\%). Otherwise, among all methods, \textbf{M6} exhibits the fastest convergence. Specifically, it converges within approximately 1300 steps, whereas the other baseline methods typically require about 3000--5000 steps for convergence.

\begin{figure*}[t]
    \centering
    \includegraphics[width=17.5cm,height=7.5cm]{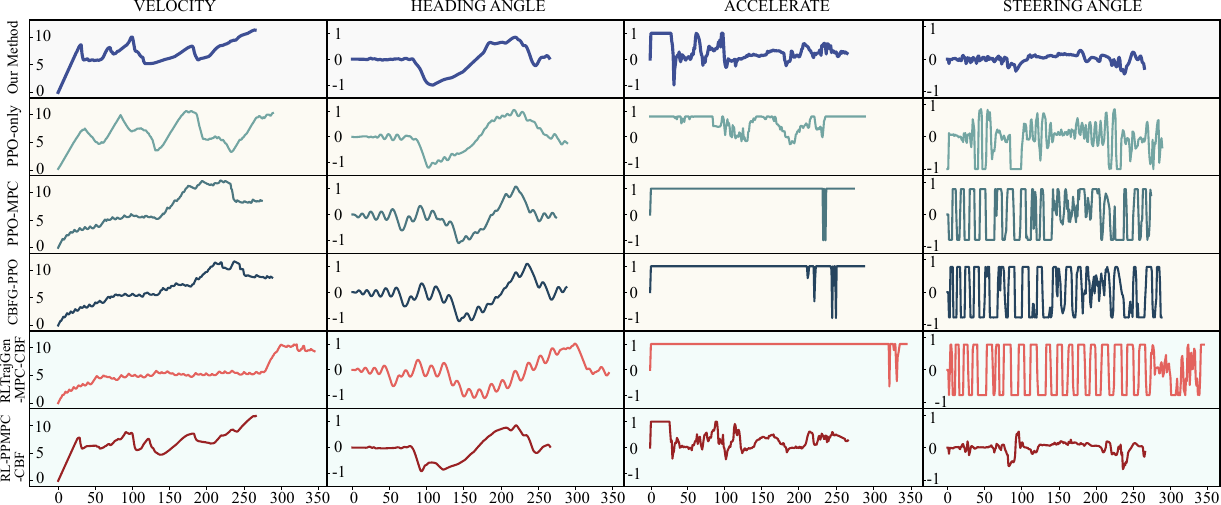}
    \caption{ Velocity, heading angle, acceleration, and steering angle profiles compared with baselines and ablation studies. }
    \vspace{-5pt}
    \label{fig:phsical-state}
\end{figure*}

\begin{figure}[t]
    \centering
    \includegraphics[width=8.5cm,height=4cm]{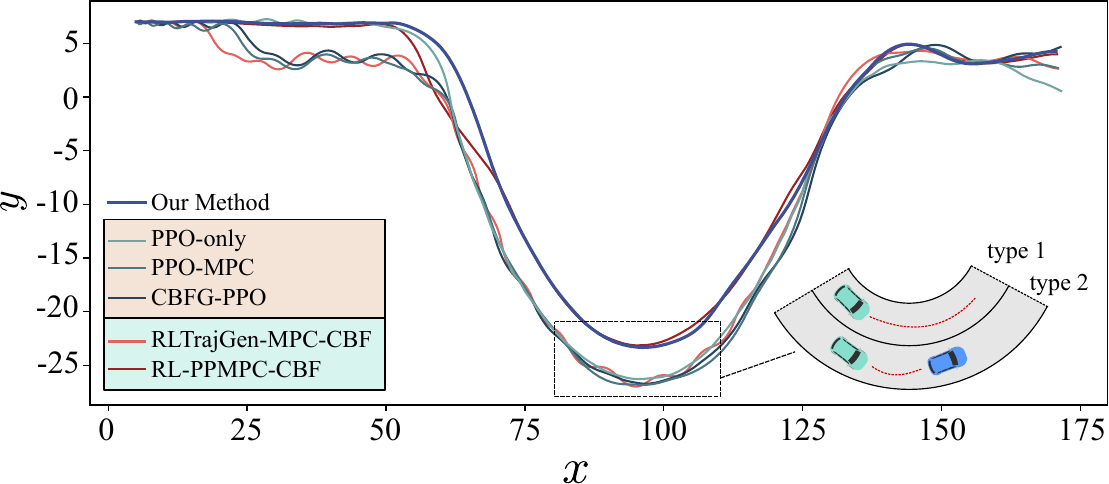}
    \caption{Visualized real-time trajectories for various methods.}
    \vspace{-5pt}
    \label{fig:position}
\end{figure}

The optimal policy from each method is implemented in the ego vehicle and assessed in the simulation environment. Fig. \ref{fig:phsical-state} illustrates the changes in velocity, heading angle, acceleration, and steering angle during driving, while Fig. \ref{fig:position} displays the actual driven trajectory. Compared to the baselines, our method keeps acceleration and steering angle within a narrower range, ensuring smoother and safer vehicle behavior. Additionally, our approach employs a distinct route planning strategy by selecting lanes with fewer surrounding vehicles and executing gradual lane changes at lower speeds, thereby enhancing ride comfort in autonomous scenarios.

\subsection{Ride comfort and behavior} \label{md:comfort}
To evaluate the comfort of algorithms during driving, appropriate metrics must be selected. Following the ISO 2631 standard, the weighted root-mean-square (RMS) of each acceleration component is commonly used to measure vehicle roughness~\cite{aledhariMotionComfortOptimizationforAutonomous2024}. Accordingly, the motion sickness dose value (MSDV) is adopted as an indicator of motion sickness and is defined as:
\begin{equation}
\mathrm{MSDV} = \sqrt{\int_0^T [\tilde{a}(t)]^2 \,\mathrm{d}t}
\end{equation}
where $T$ denotes the duration of observation and $\tilde{a}(t)$ represents the weighted acceleration (including both longitudinal and lateral components). Given both longitudinal and lateral accelerations, the discrete forms of MSDV are presented as:
\begin{equation}
\mathrm{MSDV}_V = \sqrt{\frac{1}{N}\sum_{i=1}^N a_{x,i}^2},\, \mathrm{MSDV}_L = \sqrt{\frac{1}{N}\sum_{i=1}^N a_{y,i}^2}
\end{equation}
where $a_{x,i}$ is the longitudinal acceleration at time step $i$, directly obtained from the simulation. $a_{y,i}$ is the lateral acceleration at time step $i$, estimated by $a_{y,i} = (\delta_{i+1} - \delta_i)v_i / \Delta t$.

\begin{figure}[t]
    \centering
    \includegraphics[width=8.5cm,height=7.5cm]{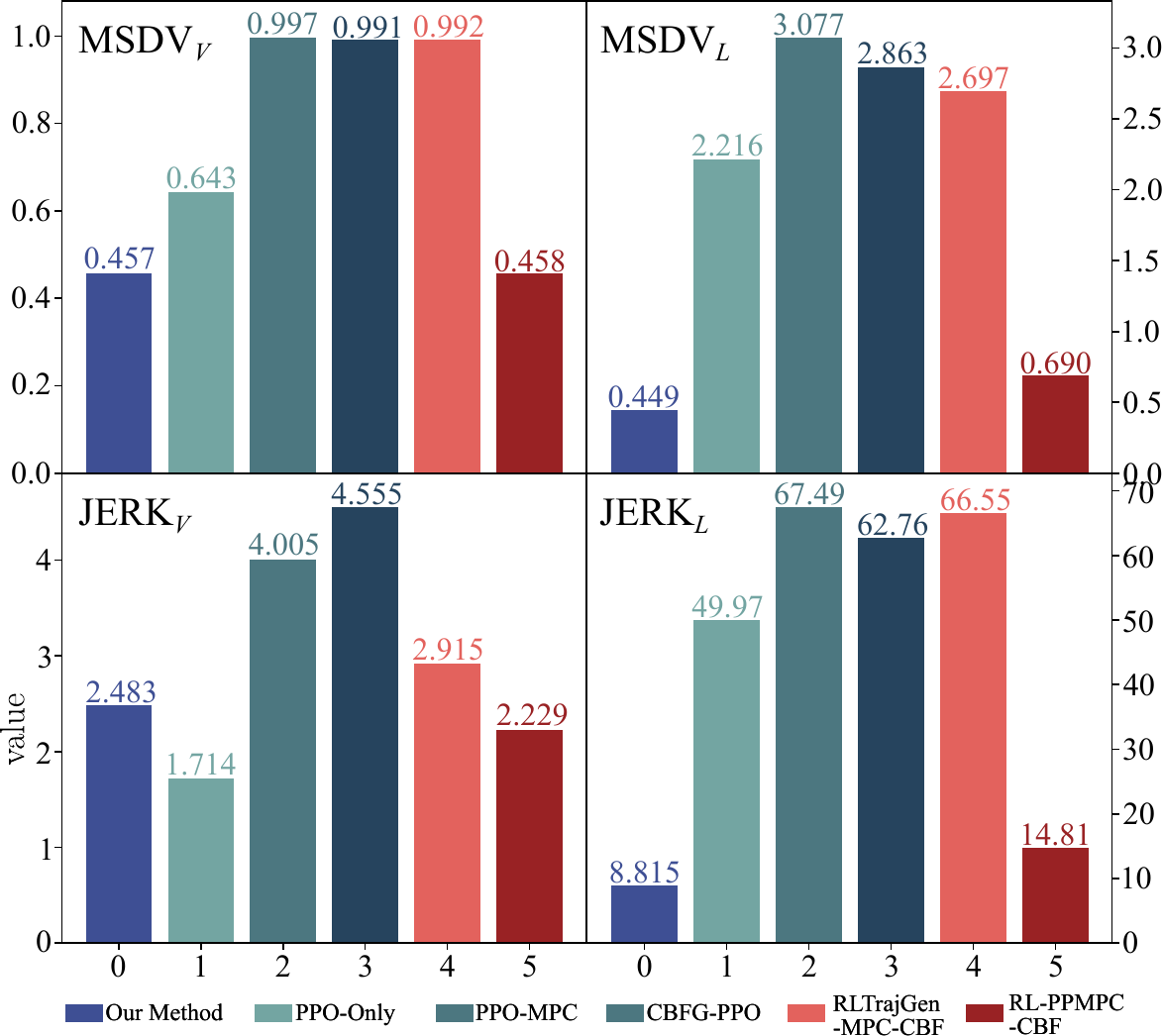}
    \caption{Comparison of $\mathrm{MSDV}$ and $\mathrm{JERK}$.}
    \vspace{-15pt}
    \label{fig:msdv-jerk-bar}
\end{figure}

\begin{table}[tp]
    \centering
    \caption{Indicator Comparison}
    \resizebox{\linewidth}{!}{%
        \begin{tabular}{c c c c c c c}
\toprule
Method      & Avg. Rwd     & Avg. Succ. Rate      & $\mathrm{MSDV}_V$  & $\mathrm{MSDV}_L$ & $\mathrm{JERK}_V$ & $\mathrm{JERK}_L$  \\
\midrule
\textbf{M1} & 156.74 & 
67.50\% & 
0.643 & 
2.216 &  
\colorbox{yellow!25}{1.714}    
&  49.97    \\

\textbf{M2} & 159.18  & 
63.75\%  & 
\colorbox{red!25}{0.997}  & 
\colorbox{red!25}{3.077}  &
4.005  &
\colorbox{red!25}{67.49}    \\

\textbf{M3} & 161.59  &
77.50\%  &
0.991    &
2.863    &
\colorbox{red!25}{4.555}  &  
62.76    \\

\textbf{M4} & \colorbox{red!25}{138.68}  & 
\colorbox{red!25}{38.75\%}  &
0.992  &
2.697  &
2.915  & 
66.55  \\

\textbf{M5} & 170.02  &
90.00\%  &
0.458    & 
2.229    &
2.229    &
14.81    \\

\textbf{M6} & \colorbox{yellow!25}{178.05}   &
\colorbox{yellow!25}{96.25\%}  & 
\colorbox{yellow!25}{0.457}    & 
\colorbox{yellow!25}{0.690}    &
2.483   &
\colorbox{yellow!25}{8.815} \\

\bottomrule
\end{tabular}
    }
    \vspace{-15pt}
    \label{tab:indicator-comparison}
\end{table}

In addition to MSDV, human perception is particularly sensitive to changes in acceleration, motivating the use of JERK as another comfort metric, defined as:
\begin{equation}
\mathrm{JERK} = \sqrt{\int_0^T \left[\frac{\mathrm{d}a}{\mathrm{d}t}\right]^2 \mathrm{d}t}
\end{equation}
Its discrete longitudinal and lateral forms are given by:
\begin{equation}
\mathrm{JERK}_V = \sqrt{\sum_{i=1}^N \mathrm{diff}(a_{x,i})^2},\, \mathrm{JERK}_L = \sqrt{\sum_{i=1}^N \mathrm{diff}(a_{y,i})^2}
\end{equation}

All methods in Table~\ref{tab:methods} are evaluated using the MSDV and JERK metrics, and the corresponding results are illustrated in Fig. \ref{fig:msdv-jerk-bar}. \textbf{M6} achieves the lowest values in $\mathrm{MSDV}_V$ (0.457), $\mathrm{MSDV}_L$ (0.449), and $\mathrm{JERK}_L$ (8.815). Although its $\mathrm{JERK}_V$ value (2.483) is slightly higher than that of \textbf{M1} (1.714) and \textbf{M5} (2.229), the proposed method overall demonstrates the best ride comfort among all evaluated approaches.

\subsection{Ablation Study and Analysis}
To validate the effectiveness of RLTrajGen and PPMPC, we conducted two ablation experiments: RLTrajGen-MPC-CBF and RL-PPMPC-CBF. These experiments are designed to explore the individual contributions of each module. Detailed descriptions of the methods can be found in Table~\ref{tab:methods}.


\subsubsection{Effect of the RLTrajGen module}
The replacement of the RL module in \textbf{M3} with the RLTrajGen module results in \textbf{M4}. In terms of AR, \textbf{M4} demonstrates faster convergence compared to \textbf{M3}, indicating that RLTrajGen significantly enhances convergence speed. However, the overall performance of \textbf{M4} is notably inferior to that of other benchmarks, suggesting that merely incorporating RLTrajGen is not sufficient for optimal results. In addition, \textbf{M5} removes the RLTrajGen module from the complete framework \textbf{M6}. The experimental results show that \textbf{M6} achieves faster convergence in both AR and ASR than \textbf{M5}. These findings indicate that the Trajectory Generator resolves the structural mismatch between RL and MPC which accelerated the agent's learning process.


\subsubsection{Effect of the PPMPC module}
\textbf{M5} is obtained by replacing the MPC module in \textbf{M3} with PPMPC. This leads to a substantial change in vehicle behavior: the overall profiles of velocity, heading angle, acceleration, and steering angle reduce to fewer oscillations, resulting in a smoother trajectory in Fig. \ref{fig:position}. This improvement arises because traditional MPC often neglects the cost induced by trajectory tracking errors, focusing instead on reducing control effort, which can gradually deteriorate performance. Meanwhile, constraining trajectory tracking errors with fixed weights is generally impractical: if the weight is set too small, MPC may ignore trajectory errors, whereas if it is set too large, the optimization problem may become infeasible. Therefore, the PPMPC module needs to be integrated with RL for more flexible parameters. By comparing the average success rate of \textbf{M5} and \textbf{M3}, PPMPC substantially improves safety, increasing the success rate from 77.50\% to 96.25\%. In addition, PPMPC achieves markedly better performance in terms of MSDV and JERK than \textbf{M3}, further demonstrating enhanced ride comfort.



\section{Conclusion}\label{sec:conclusion}




This work proposes a hierarchical learning based control framework for roundabout navigation of CAVs. Our approach tackles the oscillation issues caused by the structural mismatch between RL and MPC to facilitate smoother and more gradual decision-making. Trajectories are generated from RL policies and utilized to solve an MPC problem with PPC defined costs. Safety is ensured by a CBF layer, allowing the agent to learn the revised actions. Simulations on the MetaDrive platform validate the superior performance of our algorithm through the construction and modular comparison of five baseline methods. Our framework not only achieves the highest average reward and success rate, but also returns the smoothest trajectory measured by MSDV and JERK, demonstrating its potential for real-world driving scenarios. Future research will extend this framework to enhance navigation robustness in more complex dynamic environments.


\bibliographystyle{ieeetr}
\bibliography{utils/ref}

@inproceedings{agrawalDiscreteControlBarrier2017,
  title={Discrete control barrier functions for safety-critical control of discrete systems with application to bipedal robot navigation},
  author={Agrawal, Ayush and Sreenath, Koushil},
  booktitle={Robotics: Sci. and Sys.},
  volume={13},
   pages={1--10},
  year={2017}
}

@article{ahmedTechnologyDevelopmentsImpacts2022,
  title={Technology developments and impacts of connected and autonomous vehicles: An overview},
  author={Ahmed, Hafiz Usman and Huang, Ying and Lu, Pan and Bridgelall, Raj},
  journal={Smart Cities},
  volume={5},
  number={1},
  pages={382--404},
  year={2022},
  publisher={MDPI}
}

@article{aledhariMotionComfortOptimizationforAutonomous2024,
  title={Motion comfort optimization for autonomous vehicles: Concepts, methods, and techniques},
  author={Aledhari, Mohammed and Rahouti, Mohamed and Qadir, Junaid and Qolomany, Basheer and Guizani, Mohsen and Al-Fuqaha, Ala},
  journal={IEEE Internet of Things Journal},
  volume={11},
  number={1},
  pages={378--402},
  year={2023},
  publisher={IEEE}
}

@inproceedings{amesControlBarrierFunctions2019,
  title={Control barrier functions: Theory and applications},
  author={Ames, Aaron D and Coogan, Samuel and Egerstedt, Magnus and Notomista, Gennaro and Sreenath, Koushil and Tabuada, Paulo},
  booktitle={18th European control conference},
  pages={3420--3431},
  year={2019}
}

@article{antonioMultiAgentDeepReinforcement2022,
  title={Multi-agent deep reinforcement learning to manage connected autonomous vehicles at tomorrow's intersections},
  author={Antonio, Guillen-Perez and Maria-Dolores, Cano},
  journal={IEEE Transactions on Vehicular Technology},
  volume={71},
  number={7},
  pages={7033--7043},
  year={2022},
  publisher={IEEE}
}

@article{brunkeSafeLearningRobotics2022,
  title={Safe learning in robotics: From learning-based control to safe reinforcement learning},
  author={Brunke, Lukas and Greeff, Melissa and Hall, Adam W and Yuan, Zhaocong and Zhou, Siqi and Panerati, Jacopo and Schoellig, Angela P},
  journal={Annual Review of Control, Robotics, and Autonomous Systems},
  volume={5},
  number={1},
  pages={411--444},
  year={2022},
  publisher={Annual Reviews}
}

@article{buDiscreteTimePrescribedPerformance2023,
  title={Discrete-time prescribed performance control of an air-vehicle’s seeker stabilized platform},
  author={Bu, Xiangwei and Liu, Zongcheng},
  journal={IEEE J. Miniaturization Air Space Syst.},
  volume={4},
  number={3},
  pages={267--273},
  year={2023},
  publisher={IEEE}
}

@article{buPrescribedPerformanceControl2023,
  title={Prescribed performance control approaches, applications and challenges: A comprehensive survey},
  author={Bu, Xiangwei},
  journal={Asian Journal of Control},
  volume={25},
  number={1},
  pages={241--261},
  year={2023},
  publisher={Wiley Online Library}
}

@article{dingAdaptiveMemoryEventTriggered2024,
  title={Adaptive memory event-triggered output feedback finite-time lane-keeping control for autonomous heavy truck with roll prevention},
  author={Ding, Fei and Zhu, Kaicheng and Liu, Jie and Peng, Chen and Wang, Yafei and Lu, Junguo},
  journal={IEEE Transactions on Fuzzy Systems},
  volume={32},
  number={12},
  pages={6607--6621},
  year={2024},
  publisher={IEEE}
}

@article{dingIntelligentEventTriggered2025,
  title={Intelligent event triggered lane keeping security control for autonomous vehicle under DoS attacks},
  author={Ding, Fei and Liu, Zuoyu and Wang, Yafei and Liu, Jie and Wei, Chongfeng and Nguyen, Anh-Tu and Wang, Ningsha},
  journal={IEEE Transactions on Fuzzy Systems},
  year={2025},
  publisher={IEEE},
volume = {33},
  number = {10},
  pages = {3595--3608},
  issn = {1941-0034},
}

@inproceedings{heAdaptiveCruiseControl2019,
  title={Adaptive cruise control strategies implemented on experimental vehicles: A review},
  author={He, Yinglong and Ciuffo, Biagio and Zhou, Quan and Makridis, Michail and Mattas, Konstantinos and Li, Ji and Li, Ziyang and Yan, Fuwu and Xu, Hongming},
booktitle={9th IFAC Symposium on Advances in Automotive Control},
  pages={21--27},
  year={2019}
}

@article{huCoMoCAVsCohesiveDecisionGuided2025a,
  title={CoMoCAVs: Cohesive Decision-Guided Motion Planning for Connected and Autonomous Vehicles with Multi-Policy Reinforcement Learning},
  author={Hu, Pan},
  journal={arXiv preprint arXiv:2507.14903},
  year={2025}
}

@article{liMetaDriveComposingDiverse2022,
  title={Metadrive: Composing diverse driving scenarios for generalizable reinforcement learning},
  author={Li, Quanyi and Peng, Zhenghao and Feng, Lan and Zhang, Qihang and Xue, Zhenghai and Zhou, Bolei},
  journal={IEEE transactions on pattern analysis and machine intelligence},
  volume={45},
  number={3},
  pages={3461--3475},
  year={2022},
  publisher={IEEE}
}

@article{liuCuriositydrivenReinforcementLearning2025,
  title={Curiosity-driven reinforcement learning with graph transformers for decision-making in connected and autonomous vehicles},
  author={Liu, Qi and Tang, Yujie and Li, Xueyuan and Wang, Kaifeng and Yang, Fan and Li, Zirui},
  journal={Transportation Research Part C: Emerging Technologies},
  volume={177},
  pages={105183},
  year={2025},
  publisher={Elsevier}
}

@article{liuPrescribedPerformanceModelFree2019,
  title={Prescribed performance model-free adaptive integral sliding mode control for discrete-time nonlinear systems},
  author={Liu, Dong and Yang, Guang-Hong},
  journal={IEEE Transactions on Neural Networks and Learning Systems},
  volume={30},
  number={7},
  pages={2222--2230},
  year={2018},
  publisher={IEEE}
}

@inproceedings{lubarsCombiningReinforcementLearning2021,
  title={Combining reinforcement learning with model predictive control for on-ramp merging},
  author={Lubars, Joseph and Gupta, Harsh and Chinchali, Sandeep and Li, Liyun and Raja, Adnan and Srikant, R and Wu, Xinzhou},
  booktitle={IEEE International Intelligent Transportation Systems Conference},
  pages={942--947},
  year={2021}
}

@article{nakayaExplorationDeepReinforcement2025,
  title={Exploration of deep reinforcement learning method for cooperative control of connected automated vehicles at a roundabout},
  author={Nakaya, Reo and Harada, Tomohiro and Miura, Yukiya and Hattori, Kiyohiko and Matsuoka, Johei},
  journal={Artificial Life and Robotics},
  pages={1--17},
  year={2025},
  publisher={Springer}
}

@article{oliveiradasilvaIntegratingReinforcementLearning2025a,
  title={Integrating Reinforcement Learning and Model Predictive Control for Mixed-Logical Dynamical Systems},
  author={da Silva, Caio Fabio Oliveira and Dabiri, Azita and De Schutter, Bart},
  journal={IEEE Open Journal of Control Systems},
  year={2025},
  publisher={IEEE},
  volume = {4},
  pages = {316--331},
  issn = {2694-085X}
}

@inproceedings{pachpolEnhancedLaneKeeping2024,
  title={Enhanced Lane Keeping Assistance through IoT Sensor Fusion and Deep Learning},
  author={Pachpol, Shreyash and Vignesh, C Chandru and Jeslin, J Gnana and Kumar, BV Praveen and Viswanathan, C and Srinivasan, C},
  booktitle={Second Intl. Conf. on Intelligent Cyber Physical Systems and Internet of Things},
  pages={419--424},
  year={2024}
}

@article{ranaConnectedAutonomousVehicles2023,
  title={Connected and autonomous vehicles and infrastructures: A literature review},
  author={Rana, Md Masud and Hossain, Kamal},
  journal={International Journal of Pavement Research and Technology},
  volume={16},
  number={2},
  pages={264--284},
  year={2023},
  publisher={Springer}
}

@article{reiterSynthesisModelPredictive2025a,
  title={Synthesis of model predictive control and reinforcement learning: Survey and classification},
  author={Reiter, Rudolf and Hoffmann, Jasper and Reinhardt, Dirk and Messerer, Florian and Baumg{\"a}rtner, Katrin and Sawant, Shambhuraj and Boedecker, Joschka and Diehl, Moritz and Gros, Sebastien},
  journal={Annual Reviews in Control},
  volume={61},
  pages={101045},
  year={2026},
  publisher={Elsevier}
}

@article{seoSafetyCriticalControlNonaffine2022,
  title={Safety-critical control with nonaffine control inputs via a relaxed control barrier function for an autonomous vehicle},
  author={Seo, Joohwan and Lee, Joonho and Baek, Eunkyu and Horowitz, Roberto and Choi, Jongeun},
  journal={IEEE Robotics and Automation Letters},
  volume={7},
  number={2},
  pages={1944--1951},
  year={2022},
  publisher={IEEE}
}

@inproceedings{wangResearchAutonomousDriving2024,
  title={Research on autonomous driving decision-making strategies based deep reinforcement learning},
  author={Wang, Zixiang and Yan, Hao and Wei, Changsong and Wang, Junyu and Bo, Shi and Xiao, Minheng},
  booktitle={Proc. of the 4th International conference on internet of things and machine learning},
  pages={211--215},
  year={2024}
}

@article{wuRecentAdvancesReinforcement2024,
  title={Recent advances in reinforcement learning-based autonomous driving behavior planning: A survey},
  author={Wu, Jingda and Huang, Chao and Huang, Hailong and Lv, Chen and Wang, Yuntong and Wang, Fei-Yue},
  journal={Transportation Research Part C: Emerging Technologies},
  volume={164},
  pages={104654},
  year={2024},
  publisher={Elsevier}
}

@article{xiaoComprehensiveReviewDevelopment2010,
  title={A comprehensive review of the development of adaptive cruise control systems},
  author={Xiao, Lingyun and Gao, Feng},
  journal={Vehicle system dynamics},
  volume={48},
  number={10},
  pages={1167--1192},
  year={2010},
  publisher={Taylor \& Francis}
}

@book{xiaoSafeAutonomyControl2023,
  title={Safe autonomy with control barrier functions: Theory and applications},
  author={Xiao, Wei and Cassandras, Christos G and Belta, Calin},
  year={2023},
  publisher={Springer}
}

@article{zhangControlBarrierFunctionGuided2025,
  title={Control barrier function-guided deep reinforcement learning for decision-making of autonomous vehicle at on-ramp merging},
  author={Zhang, Changzhu and Dai, Lifei and Zhang, Hao and Wang, Zhuping},
  journal={IEEE Transactions on Intelligent Transportation Systems},
  year={2025},
  publisher={IEEE},
  volume = {26},
  number = {6},
  pages = {8919--8932},
  issn = {1524-9050, 1558-0016},
}

@article{zhouDevelopmentEfficientDriving2020,
  title={Development of an efficient driving strategy for connected and automated vehicles at signalized intersections: A reinforcement learning approach},
  author={Zhou, Mofan and Yu, Yang and Qu, Xiaobo},
  journal={IEEE Transactions on Intelligent Transportation Systems},
  volume={21},
  number={1},
  pages={433--443},
  year={2019},
  publisher={IEEE}
}

@inproceedings{zhouEnhancePlanningPhysicsinformed2024,
  title={Enhance planning with physics-informed safety controller for end-to-end autonomous driving},
  author={Zhou, Hang and Liu, Haichao and Lu, Hongliang and Ma, Jun and Ji, Yiding},
  booktitle={IEEE Intl. Conf. on Rob. and Bio.},
  pages={1775--1782},
  year={2024}
}

@article{zhang2026safety,
  title={Safety-critical Control under Timed Reach-Avoid Specifications: A Backup Control Barrier Function Approach},
  author={Zhang, Yangrui and Jiang, Tianjian and Cai, Zhongxuan and Ji, Yiding},
  journal={IEEE Control Sys. Letters},
  volume={10},
  pages={109 - 114},
  year={2026}
}

@inproceedings{huang2025differentiable,
  title={A Differentiable Constraint-Aware Motion Planning Module for Unsupervised Trajectory Optimization},
  author={Huang, Zhaoyan and Zhang, Chao and Zhang, Rui and Ji, Yiding and Hong, Jinlong and Gao, Bingzhao and Liu, Qingwei},
  booktitle={IEEE 28th International Conference on Intelligent Transportation Systems},
  pages={1294--1300},
  year={2025}
}

@inproceedings{zhou2026physics,
  title={Physics-informed Diffusion {M}amba Transformer for Real-world Driving},
  author={Zhou, Hang and Zhang, Qiang and Liu, Peiran and Qin, Yihao and Yan, Zhaoxu and Ji, Yiding},
  booktitle={IEEE International Conference on Robotics and Automation},
  year={2026}
}

\end{document}